\documentclass[10pt,twocolumn,letterpaper]{article}

\makeatletter
\renewcommand{\paragraph}{%
  \@startsection{paragraph}{4}%
  {\z@}{1ex \@plus 1ex \@minus .2ex}{-1em}%
  {\normalfont\normalsize\bfseries}%
}
\makeatother

\usepackage{cvpr}
\usepackage{times}
\usepackage{epsfig}
\usepackage{graphicx}
\usepackage{amsmath}
\usepackage{amssymb}
\usepackage{mathtools}
\usepackage{subfigure}

\DeclarePairedDelimiter\abs{\lvert}{\rvert}%
\usepackage{paralist}
\usepackage{multirow}
\usepackage{makecell}
\usepackage{soul}
\usepackage{sidecap}

\usepackage[dvipsnames]{xcolor}

\newcommand{\tabincell}[2]{\begin{tabular}{@{}#1@{}}#2\end{tabular}}  
\renewcommand{\hl}[1]{#1}

\usepackage[breaklinks=true,bookmarks=false]{hyperref}

\cvprfinalcopy 

\def\cvprPaperID{4777} 

\ifcvprfinal\fi
\begin{document}

\title{
	BANet: Bidirectional Aggregation Network \\ 
with Occlusion Handling for Panoptic Segmentation
}

\author{
Yifeng Chen$^{1}$, Guangchen Lin$^{1}$, Songyuan Li$^{1}$, Bourahla Omar$^{1}$, Yiming Wu$^{1}$, \\
Fangfang Wang$^{1}$, Junyi Feng$^{1}$, Mingliang Xu$^{2}$, Xi Li$^{1}$\thanks{Corresponding author, xilizju@zju.edu.cn} \\
\\
$^1$Zhejiang University, $^2$Zhengzhou University\\
\small{\texttt{\{yifengchen, aaronlin, leizungjyun, xilizju\}@zju.edu.cn}} \\
}

\maketitle
\thispagestyle{empty}

\begin{abstract}
   Panoptic segmentation aims to perform instance segmentation for foreground instances and semantic segmentation for background stuff simultaneously. The typical top-down pipeline concentrates on two key issues: 1) \textit{how to effectively model the intrinsic interaction between semantic segmentation and instance segmentation}, and 2) \textit{how to properly handle occlusion for panoptic segmentation}. Intuitively, the complementarity between semantic segmentation and instance segmentation can be leveraged to improve the performance. Besides, we notice that using detection/mask scores is insufficient for resolving the occlusion problem. Motivated by these observations, we propose a novel deep panoptic segmentation scheme based on a bidirectional learning pipeline. Moreover, we introduce a plug-and-play occlusion handling algorithm to deal with the occlusion between different object instances. The experimental results on COCO panoptic benchmark validate the effectiveness of our proposed method. 
   Codes will be released soon at \href{https://github.com/Mooonside/BANet}{https://github.com/Mooonside/BANet}.
\end{abstract}

\section{Introduction}
Panoptic segmentation~\cite{kirillov2019panoptic1}, an emerging and challenging problem in computer vision, is a composite task unifying both semantic segmentation (for background stuff) and instance segmentation (for foreground instances).
A typical solution to the task is in a top-down deep learning mannerwhereby instances are first identified and then assigned to semantic labels~\cite{Li2018Learning,  li2019attention, liu2019end, xiong2019upsnet}. In this way, two key issues arise out of a robust solution:
\begin{inparaenum}[1)]
	\item how to effectively model the intrinsic interaction between semantic segmentation and instance segmentation, and
	\item how to robustly handle the occlusion for panoptic segmentation.
\end{inparaenum}

\begin{figure}[tb]
	\centering
	\includegraphics[width=1.0\linewidth]{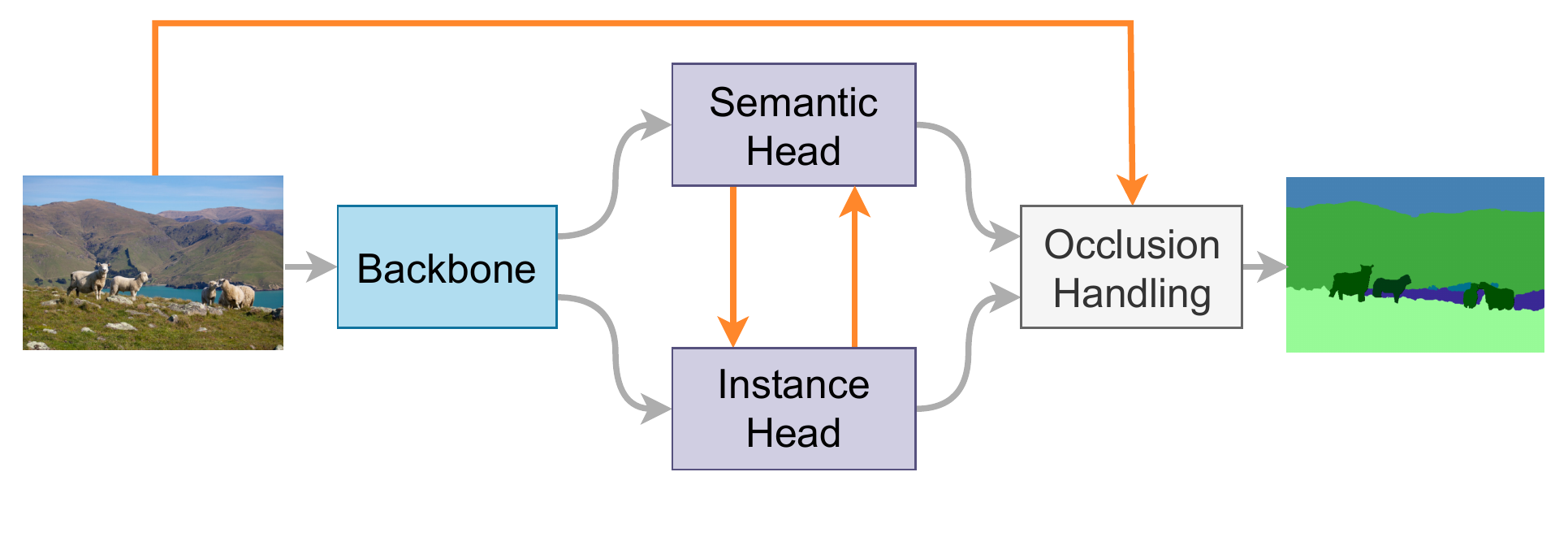}
	\caption{\footnotesize The illustration of BANet. We introduce a bidirectional path to leverage the complementarity between semantic and instance segmentation. To obtain the panoptic segmentation results, low-level appearance information is utilized in the occlusion handling algorithm.}
\label{fig:Pipeline_b}
\end{figure}

In principle, the complementarity does exist between the tasks of semantic segmentation and instance segmentation. 
Semantic segmentation concentrates on capturing the rich pixel-wise class information for 
scene understanding. Such information could work as useful contextual clues to enrich the features for instance segmentation. Conversely, instance segmentation gives rise to the structural information (e.g., shape) on object instances,
which enhances the discriminative power of the feature representation for semantic segmentation. Hence, the interaction between
these two tasks is bidirectionally reinforced and reciprocal. 
However, previous works~\cite{Li2018Learning, li2019attention,xiong2019upsnet} usually take a unidirectional learning pipeline to use score maps from instance segmentation to guide semantic segmentation, resulting in the lack of a path from semantic segmentation to instance segmentation. Besides, the information contained by these instance score maps is often coarse-grained with a very limited channel size, leading to the difficulty in encoding more fine-grained structural information for semantic segmentation.

In light of the above issue, we propose a Bidirectional Aggregation NETwork, dubbed BANet, for panoptic segmentation
to model the intrinsic interaction between semantic segmentation and instance segmentation at the feature level. 
Specifically, BANet possesses bidirectional paths for feature aggregation between these two tasks, which
respectively correspond to two modules:
 Instance-To-Semantic~(I2S) and Semantic-To-Instance~(S2I). 
S2I passes the context-abundant 
features from semantic segmentation to instance segmentation for localization and recognition. 
Meanwhile, the instance-relevant features, attached with more structural information, are fed back to semantic segmentation to enhance the discriminative capability of the semantic features. To achieve a precise instance-to-semantic feature transformation, we design the ROIInlay operator based on bilinear interpolation. 
This operator is capable of restoring the structure of cropped instance features so that they can be aggregated with the semantic features for semantic segmentation.

After the procedures of semantic and instance segmentation, we need to fuse their results into
the panoptic format. During this fusion process, a key problem is to reason the occlusion relationships
for the occluded parts among object instances. A conventional way~\cite{de2018panoptic, kirillov2019panoptic1, liu2019end, xiong2019upsnet}
relies heavily on detection/mask scores, which are often inconsistent with the actual spatial ranking relationships of object instances.
For example, a tie usually overlaps a person, but it tends to get a lower score~(due to class imbalance).
With this motivation, we propose a learning-free occlusion handling algorithm based on 
the affinity between the overlapped part and each object instance in
the low-level appearance feature space. It compares the similarity between occluded parts and object instances and assigns each part to the object of the closest appearance.

In summary, the contributions of this work are as follows:
\begin{itemize}
	\item We propose a deep panoptic segmentation scheme based on a bidirectional learning pipeline, namely Instance-To-Semantic~(I2S) and Semantic-To-Instanc-e~(S2I) to enable feature-level interaction between instance segmentation and semantic segmentation.
	\item We present the  ROIInlay operator to achieve the precise instance-to-semantic feature mapping
          from the cropped bounding boxes to the holistic scene image. 
	\item We propose a simple yet effective learning-free approach to handle the occlusion, which can be plugged in any top-down based network.
\end{itemize}

\section{Related Work}
\paragraph{Semantic segmentation} Semantic segmentation, the task of assigning a semantic category to each pixel in an image, has made great progress recently with the development of the deep CNNs in a fully convolutional fashion (FCN\cite{fcn}). It has been known that contextual information is beneficial for segmentation~\cite{dai2015convolutional,farabet2013learning,gould2009decomposing,he2004multiscale,kohli2008robust,ladicky2009associative,mostajabi2015feedforward,shotton2009textonboost}, and these models usually provide a mechanism to exploit it. For example, PSPNet~\cite{zhao2017pyramid} features global pyramid pooling which provides additional contextual information to FCN. Feature Pyramid Network (FPN)~\cite{lin2017feature}  takes features from different layers as multi-scale information and stacks them to a feature pyramid. DeepLab series~\cite{chen2018deeplab, chen2018encoder} apply several architectures with atrous convolution to capture multi-scale context. In our work, we focus on utilizing features from semantic segmentation to help instance segmentation instead of designing a sophisticated context mechanism.

\paragraph{Instance segmentation}
Instance segmentation assigns a category and an instance identity to each object pixel in an image.  Methods for instance segmentation fall into two main categories: top-down and bottom-up. The top-down, or proposal-based, methods~\cite{chen2018masklab, dai2016instance,dai2017deformable,he2017mask,li2017fully,liu2018path,peng2018megdet} first generate bounding boxes for object detection, and then perform dense prediction for instance segmentation. The bottom-up, or segmentation-based, methods~\cite{bai2017deep, kendall2018multi, fathi2017semantic,liang2017proposal,liu2017sgn,liu2018affinity,newell2017associative,uhrig2016pixel,zhang2016instance,zhang2015monocular} first perform pixel-wise semantic segmentation, and then extract instances out of grouping. Top-down approaches dominates the leaderboards of instance segmentation. We adopt this manner for the instance segmentation branch in our pipeline. Chen~\etal~\cite{chen2019hybrid} made use of semantic features in instance segmentation. Our approach is different from it in that we design a bidirectional path between instance segmentation and semantic segmentation.


\paragraph{Panoptic segmentation} 
Panoptic segmentation unifies semantic and instance segmentation, and therefore its methods can also fall into top-down and bottom-up categories on the basis of their strategy to do instance segmentation. Kirillov~\etal  \cite{kirillov2019panoptic1} proposed a baseline that combines the outputs from Mask-RCNN~\cite{he2017mask} and PSPNet~\cite{zhao2017pyramid} by heuristic fusion.  
De Geus~\etal~\cite{de2018panoptic} and Kirillov~\etal~\cite{kirillov2019panoptic2} proposed end-to-end networks with multiple heads for panoptic segmentation. To model the internal relationship between instance segmentation and semantic segmentation, previous works~\cite{Li2018Learning, li2019attention} utilized class-agnostic score maps to guide semantic segmentation. 

To solve occlusion between objects, Liu \etal~\cite{liu2019end} proposed a spatial ranking module to predict the ranking of objects and Xiong~\etal~\cite{xiong2019upsnet} proposed a parameter-free module to bring explicit competition between object scores and semantic logits.

 Our approach is different from previous works in three ways. 
\begin{inparaenum}[1)]
	\item We utilize instance features instead of coarse-grained score maps to improve the discriminative ability of semantic features.
	\item We build a path from semantic segmentation to instance segmentation.
	\item We make use of low-level appearance to resolve occlusion.
\end{inparaenum}

\section{Methods}
\begin{figure*}[t]
	\centering
	\includegraphics[clip,width=0.95\textwidth]{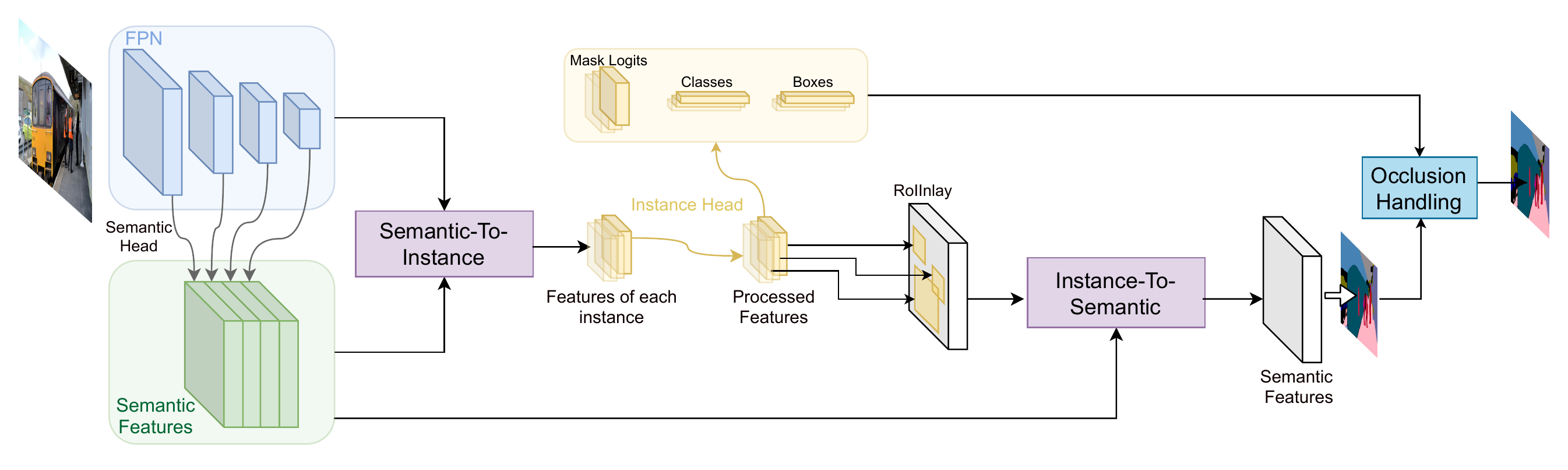}
		\caption{\footnotesize Our framework takes advantage of complementarity between semantic and instance segmentation. This is shown through two key modules, namely, Semantic-To-Instance~(S2I) and Instance-To-Semantic~(I2S). S2I uses semantic features to enhance instance features.
I2S uses instance features restored by the proposed RoIInlay operation for better semantic segmentation. 
After performing instance and semantic segmentation, the occlusion handling module is applied to determine the belonging of occluded pixels and merge the instance and semantic outputs as the final panoptic segmentation.
		}
	\label{fig:arch}
\end{figure*}


Our BANet contains four major components: a backbone network, the Semantic-To-Instance~(S2I) module, the Instance-To-Semantic~(I2S) module and an occlusion handling module, as shown in Figure~\ref{fig:arch}, We adopt ResNet-FPN as the backbone.
The S2I module aims to use semantic features to help instance segmentation as described in Section~\ref{sec:s2i}. The I2S module assists semantic segmentation with instance features as described in Section~\ref{sec:i2s}.  In Section~\ref{sec::texture}, an occlusion handling algorithm is proposed to deal with instance occlusion.


\subsection{Instance Segmentation}
\label{sec:s2i}
\begin{figure}[tb]
	\centering
	\includegraphics[width=\linewidth]{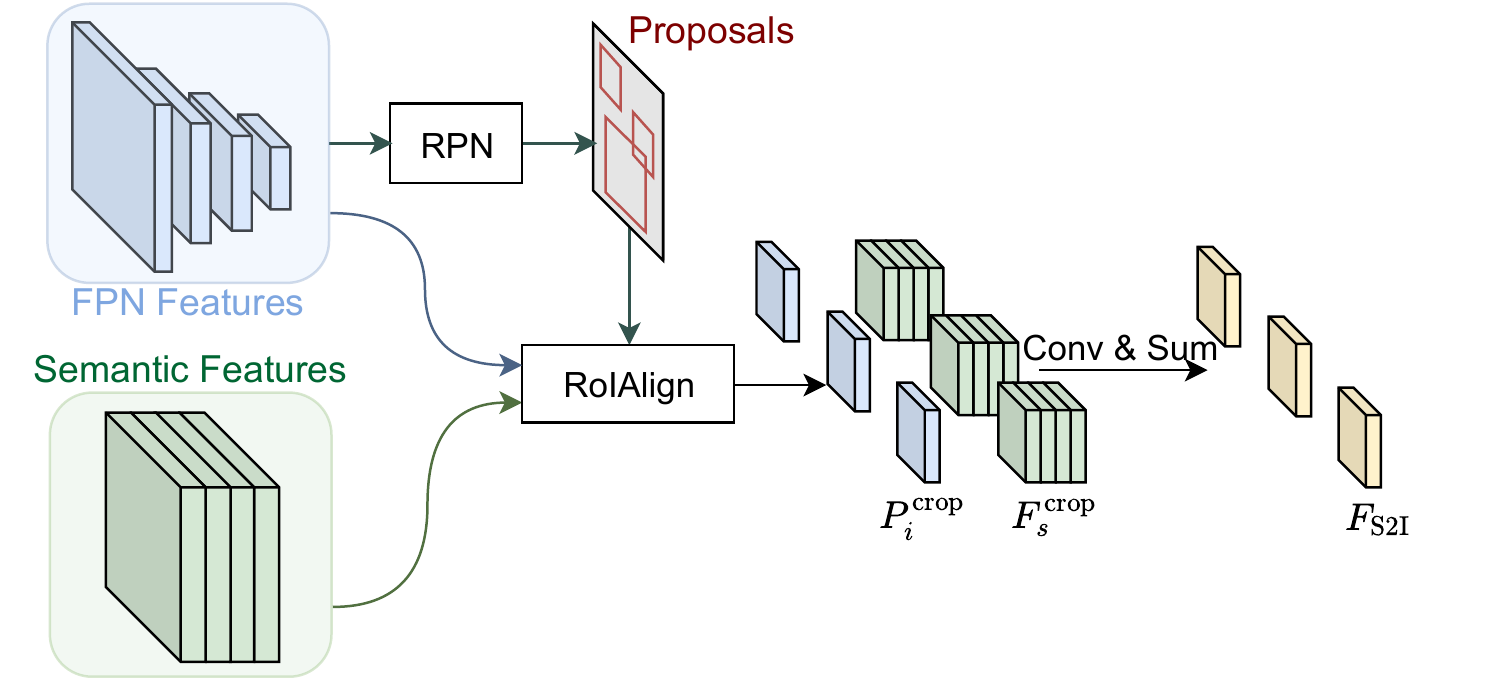}

	\caption{\footnotesize The architecture of our S2I module. For each instance, S2I crops semantic features and the selected FPN features of the instance and then aggregates the cropped features. As a result, it enhances instance segmentation by semantic information.}


	\label{fig:S2I}
\end{figure}

Instance segmentation is the task of localizing, classifying and predicting a pixel-wise mask for each instance. 
We propose the S2I module to bring about contextual clues for the benefit of instance segmentation, as illustrated in Figure~\ref{fig:S2I}. The semantic features $F_S$ are obtained by applying a regular semantic segmentation head on the FPN features $\{P_i\}_{i=2...5}$.

For each instance proposal, we crop semantic features $F_S$ and the selected FPN features $P_i$ by RoIAlign~\cite{he2017mask}. These features are denoted by $F_S^{\text{crop}}$ and $P_i^{\text{crop}}$. The proposals we use here are obtained by feeding FPN features into a regular RPN head. 

After that,  $F_S^{\text{crop}}$ and ${P_i}^{\text{crop}}$  are aggregated  as follows:
\begin{equation}
	F_{\text{S2I}} = \phi(F_S^{\text{crop}}) + {P_i}^{\text{crop}},
\end{equation}
where $\phi$ is a $1\times1$ convolution layer to align the feature spaces. The aggregated features $F_{\text{S2I}}$ benefit from contextual information from $F_S^{\text{crop}}$ and spatial details from ${P_i}^{\text{crop}}$. 

$F_{\text{S2I}}$ is fed into a regular instance segmentation head to predict masks, boxes and categories for instances. The specific design of the instance head follows~\cite{he2017mask}. For mask predictions, three $3\times3$ convolutions are applied to $F_{\text{S2I}}$ to extract instance-wise features $F_{\text{ins}}$. Then a deconvolution layer up-samples the features and predicts object-wise masks of $28\times28$. Meanwhile, fully connected layers are applied to $F_{\text{S2I}}$ to predict boxes and categories. Note that $F_{\text{ins}}$ is later used in Section~\ref{sec:i2s}.

\subsection{Semantic Segmentation}
\label{sec:i2s}
Semantic segmentation assigns each pixel with a class label.  
Our framework utilizes instance features to introduce structural information to semantic features. It does so through our I2S module which uses $F_{\text{ins}}$ from the previous section.
However, $F_{\text{ins}}$ cannot be fused with semantic feature $F_S$ directly since it is already cropped and resized. To solve this issue, we propose the RoIInlay operation, which maps  $F_{\text{ins}}$ back into a feature map $F_\text{inlay}$ with the same spatial size as $F_S$. This restores the structure of each instance, allowing us to efficiently use it in semantic segmentation.

\begin{figure}[tb]
	\centering
	\includegraphics[clip,width=\linewidth]{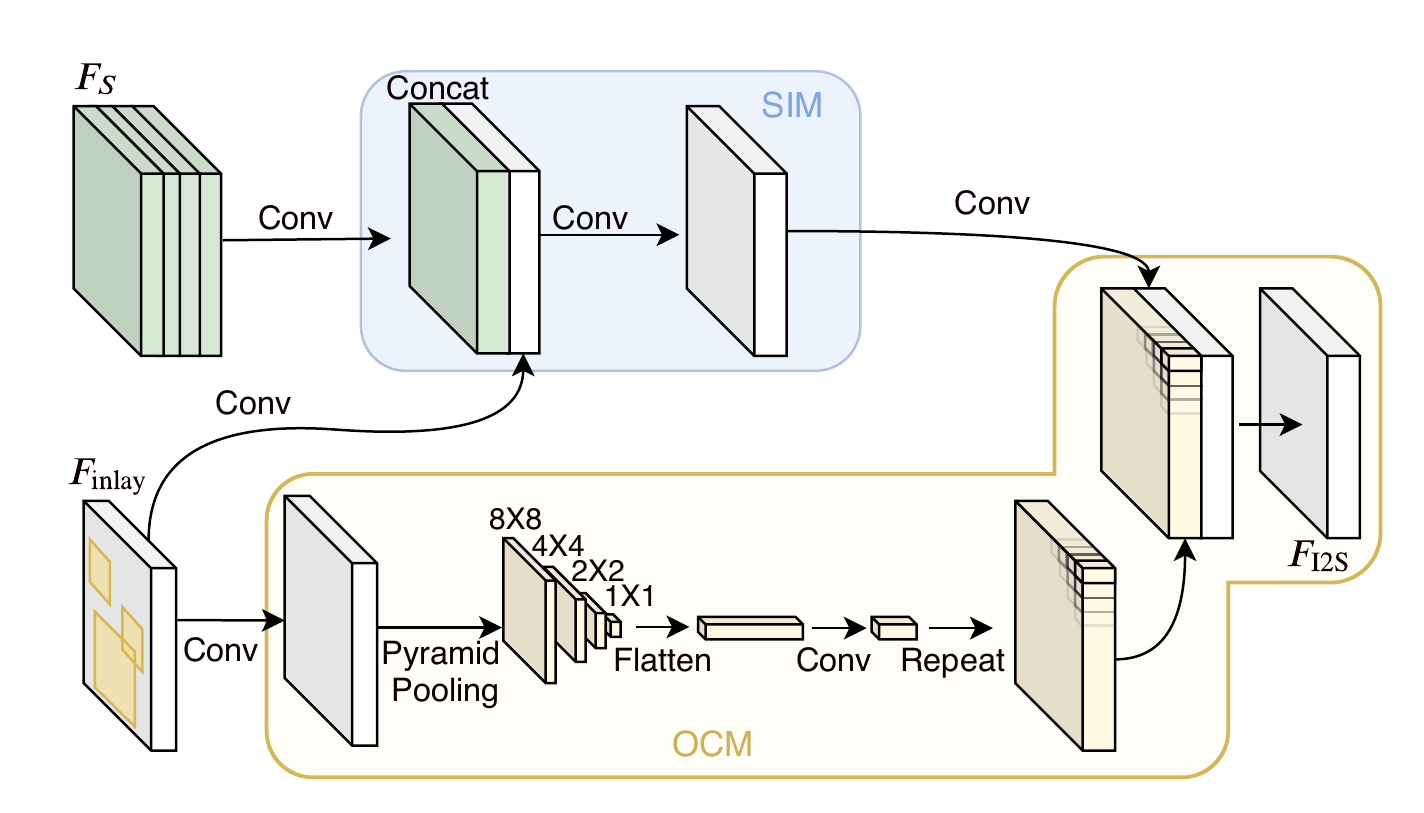}
	\caption{\footnotesize The architecture of the I2S module. SIM uses instance features restored by RoIInlay and combines them with semantic features. Meanwhile, OCM extracts information on the objects' layout in the scene. After that, OCM combines it with SIM's output for use in semantic segmentation.}

	\label{fig:i2s}
\end{figure}

After obtaining $F_{\text{inlay}}$, we use it along with $F_S$ to perform semantic segmentation. As shown in Figure~\ref{fig:i2s}, these two features are aggregated in two modules, namely Structure Injection Module~(SIM) and Object Context Module~(OCM). In SIM, $F_{\text{inlay}}$ and $F_S$ are first projected to the same feature space. Then, they are concatenated and go through a $3\times3$ convolution layer to alleviate possible distortions caused by RoIInlay. By doing so, we inject the structure information of $F_\text{inlay}$ into the semantic feature $F_S$.

OCM takes the output of SIM and further enhances it by information on the objects' layout in the scene. 

As shown in Figure\mbox{~\ref{fig:i2s}}, we first project $F_{\text{inlay}}$ into a space of $E$ dimension~($E=10$). Then, a pyramid of max-pooling is applied to get multi-scale descriptions of the objects' layout. These descriptions are flattened, concatenated and projected to obtain an encoding of the layout. 
This encoding is repeated horizontally and vertically, and concatenated with the output of SIM. Finally, the concatenated features are projected as $F_{\text{I2S}}$.

$F_{\text{I2S}}$ is then used to predict semantic segmentation
which will be later used to obtain the panoptic result. 
\paragraph{Extraction of semantic features}
To extract $F_S$, we use a semantic head with a design that follows~\cite{xiong2019upsnet}. A subnet of three stacked $3 \times 3$ convolutions is applied to each FPN feature. After that, they are upsampled and concatenated to form $F_S$. 


\paragraph{RoIInlay}
RoIInlay aims to restore features cropped by operations such as RoIAlign back to their original structure. In particular, RoIInaly resizes the cropped feature and inlays it in an empty feature map at the correct location, namely at the position from which it was first cropped. 

\begin{figure}[h]
	\centering
	\includegraphics[width=0.9\linewidth]{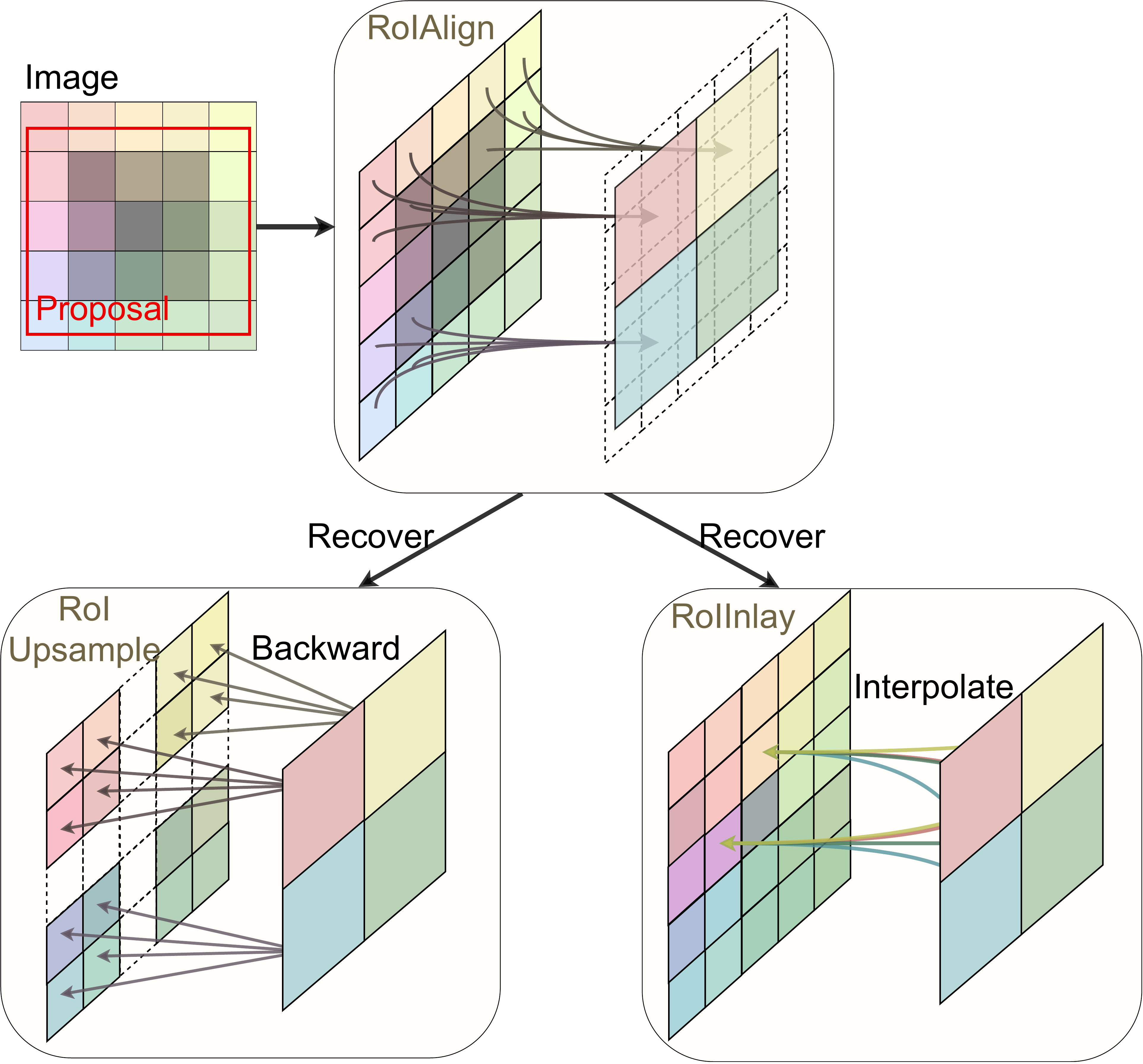}
	
	\caption{\footnotesize 
		The difference between RoIUpsample and our RoIInlay. Both RoIUpsample and RoIInlay restore features cropped by RoIAlign.
		However, RoIUpsample only uses a single reference for each pixel whereas RoIInlay uses four references and does not suffer from pixels with unassigned values.
	}
	\label{fig:RoIInlay} 
\end{figure}

As a patch-recovering operator, RoIInlay shares a common purpose with RoIUpsample\mbox{~\cite{li2019attention}}, but RoIInlay has two advantages over RoIUpsample thanks to its different interpolation style, as shown in Figure~\ref{fig:RoIInlay}.
RoIUpsample obtains values through a modified gradient function of bilinear interpolation. RoIInlay applies the bilinear interpolation carried out in the relative coordination of sampling points~(used in RoIAlign). Therefore, it can both avoid ``holes'', \mbox{\ie} pixels whose values cannot be recovered and interpolate more accurately. More comparisons on these two operators can be found in the supplementary material.

Recall that in RoIAlign\mbox{~\cite{he2017mask}}, $m\times m$ sampling points are generated to crop a region. The resulting feature is thus divided into a group of $m\times m$ bins with a sampling point at the center of each bin. Given a region of size $(w_r, h_r)$ the size of each bin will be $b_h = h_r/m$ and $b_w = w_r/m$ and the value at each sampling point is obtained by interpolating from the 4 closest pixels as shown in \mbox{Figure~\ref{fig:RoIInlay}.}


Given the positions and values of each sampling point, RoIInlay aims to recover values of pixels within the region.
To achieve this, it is designed as a bilinear interpolation carried out in the relative coordinates of sampling points. Specifically, for a pixel located at $(a, b)$, we find its four nearest sampling points $\{(x_i, y_i), i \in [1, 4]\}$. The value at $(a, b)$ is calculated as:
\begin{equation}
v(a, b) = \sum_{i=1}^{4} G(a, x_i, b_w) G(b, y_i, b_h) v(x_i, y_i),
\end{equation}
where $v(x_i,y_i)$ is the value of sampling point $(x_i,y_i)$, $(b_h, b_w)$ is the size of each sampling bin and $G$ is the bilinear interpolation kernel in the relative coordinates of sampling points:
\begin{equation}
G(a, x_i, b_w)= 1.0 - \frac{\abs{a-x_i}}{b_w}.
\end{equation}

Pixels within the region but out of the boundary of sampling points are calculated as if they were positioned at the boundary. 
To handle cases where different objects may generate values at the same position, we take the average of these values to maintain the scale.

\subsection{Occlusion Handling}
\label{sec::texture}
Occlusion occurs during instance segmentation when a pixel $x$ is claimed by multiple objects $\{O_1, \dots, O_k\}$. To get the final panoptic result, we must resolve the overlap relationships among objects so that $x$ is assigned to just one object. We argue that low-level appearance is a strong visual cue for the spatial ranking of objects compared to semantic features or instance features. The former contains mostly category information, which cannot resolve the occlusion of the objects belonging to the same class,  while the latter loses details after RoIAlign, which are fatal when small objects~(\eg tie) overlaps big ones~(\eg person).

By utilizing appearance as the reference, we propose a novel occlusion handling algorithm that assigns pixels to the most similar object instance. To compare the similarity between a pixel $x$ and an object instance $O_i$,  we need to define a measure $f(x, O_i)$. 
In this algorithm, we adopt the cosine similarity between the RGB of pixel $x$ and each object instance $O_i$~(represented by its average RGB values).

After calculating the similarity between $x$ and each object,  we assign $x$ to $O^*$, where
\begin{equation}
O^* = \text{argmax}_{O_i} f(x, O_i)
\end{equation}

In practice, instead of considering individual pixels, we consider them in sets, which will lead to more stable results.  To compare between an object and a pixel set, we average over the similarity of that object with each pixel in the set. 

Through this learning-free algorithm, the instance assignment of each pixel is resolved. After that, we combine it with the semantic segmentation for the final panoptic results according to the procedures in~\cite{kirillov2019panoptic1}.

\subsection{Training and Inference}
\paragraph{Training}
During training, we sample ground truth detection boxes and only apply RoIInlay on features of sampled objects. The sampling rate is chosen randomly from 0.6 to 1, where at least one ground truth box is kept. There are seven loss items in total. The RPN proposal head contains two losses: $L_{rpn\_cls}$ and $L_{rpn\_box}$. The instance head contains three losses: $L_{cls}$~(bbox classification loss), $L_{box}$~(bbox regress loss) and $L_{mask}$~(mask prediction loss). The semantic head contains two losses: $L_{seg}$~(semantic segmentation from $F_S$) and $L_{I2S}$~(semantic segmentation from $F_\text{I2S}$). The  total loss function $L$ is :
\begin{equation}
\begin{aligned}
L = &\underbrace{L_{rpn\_cls} + L_{rpn\_box}}_{\text{rpn proposal loss}}+\underbrace{L_{cls} + L_{box} + L_{mask}}_{\text{instance segmentation loss}}\\
&+ \underbrace{\lambda_s L_{seg} + \lambda_{i} L_{I2S}}_{\text{semantic segmentation loss}},
\end{aligned}
\end{equation}
where $\lambda_s$ and $\lambda_i$ are loss weights to control the balance between semantic segmentation and other tasks.

\paragraph{Inference}
During inference, predictions from instance head are sent to the occlusion handling module. It first performs non-maximum-suppression~(NMS) to remove duplicate predictions. Then the occluded objects are identified and their conflicts are solved based on appearance similarity. Afterwards, the occlusion-resolved instance prediction is combined with semantic segmentation prediction following~\cite{kirillov2019panoptic1}, where instances always overwrite stuff regions. Finally, stuff regions are removed and labeled as ``void'' if their areas are below a certain threshold.


\section{Experiments}
\subsection{Datasets}
We evaluate our approach on MS COCO~\cite{lin2014microsoft}, a large-scale dataset with annotations of both instance segmentation and semantic segmentation. It contains 118k training images, 5k validation images, and 20k test images. The panoptic segmentation task in COCO includes 80 thing categories and 53 stuff categories. We train our model on the \emph{train} set without extra data and report results on both \emph{val} and \emph{test-dev} sets.


\subsection{Evaluation Metrics}
\paragraph{Single-task metrics}
For semantic segmentation, the $\text{mIoU}^\text{Sf}$~(mean Intersection-over-Union averaged over stuff categories) is reported. We do not report the mIoU over thing categories since the semantic segmentation prediction of thing classes will not be used in the fusion algorithm. For instance segmentation, we report $\text{AP}_\text{mask}$, which is averaged between categories and IoU thresholds~\cite{lin2014microsoft}.  

\paragraph{Panoptic segmentation metrics}
We use PQ~\cite{kirillov2019panoptic1}~(averaged over categories) as the metric for panoptic segmentation. It captures both recognition quality~(RQ) and segmentation quality~(SQ):
\begin{equation}
\footnotesize
\mathrm{PQ}=\underbrace{\frac{\sum_{(p, g) \in T P} \operatorname{IoU}(p, g)}{|T P|}}_{\text {segmentation quality(SQ) }} \times \underbrace{\frac{|T P|}{|T P|+\frac{1}{2}|F P|+\frac{1}{2}|F N|}}_{\text {recognition quality(RQ) }},
\end{equation}
where $\operatorname{IoU}(p, g)$ is the intersection-over-union between a predicted segment $p$ and the ground truth $g$, $T P$ refers to matched pairs of segments, $F P$ denotes the unmatched  predictions and $F N$ represents the unmatched ground truth segments. Additionally, 
$\text{PQ}^{\text{Th}}$~(average over thing categories) and $\text{PQ}^\text{Sf}$~(average over stuff categories) are reported to reflect the improvement on instance and semantic segmentation segmentation. 

\begin{table*}[ht]
	\footnotesize
	\centering
	\begin{tabular}{l | c | c |  c c c | c c c |c c c}
		\Xhline{3\arrayrulewidth}
		Models & Subset	& Backbone & $\text{PQ}$ & $\text{SQ}$ & $\text{RQ}$ & 
		$\text{PQ}^\text{Th}$ & $\text{SQ}^\text{Th}$ &  $\text{RQ}^\text{Th}$& 
		$\text{PQ}^\text{Sf}$ & $\text{SQ}^\text{Sf}$ &  $\text{RQ}^\text{Sf}$ \\
		\hline
		\hline
		JSIS-Net~\cite{de2018panoptic}  & \emph{val} & ResNet-50-FPN & 26.9   &  72.4 &  35.7   &  29.3   & 72.1 & 39.2 & 23.3 & 72.0 & 30.4  \\
		Panoptic FPN~\cite{kirillov2019panoptic2} & \emph{val} & ResNet-50-FPN & 39.0   & -   &   -   &   45.9  & - & - & 28.7 & - & - \\
		OANet~\cite{liu2019end} & \emph{val}  & ResNet-50-FPN & 39.0   &   77.1  &  47.8     & 48.3  &  \textbf{81.4} & 58.0 & 24.9 & 70.6 & 32.5 \\
		AUNet~\cite{li2019attention} & \emph{val} & ResNet-50-FPN & 39.6   &  -  &  -     &  \textbf{49.1}   & - & - & 25.2 & - & -  \\
		Ours  & \emph{val}  & ResNet-50-FPN & \textbf{41.1}  &  \textbf{77.2}  & \textbf{51}  & \textbf{49.1}  & 80.4 & \textbf{60.3} & \textbf{29.1} & \textbf{72.4} & \textbf{37.1} \\
		\hline
		\hline
		$\text{UpsNet}^{\dag}$~\cite{xiong2019upsnet} & \emph{val}  & ResNet-50-FPN & 42.5   & 78.0   & 52.4      & 48.5 &79.5 & 59.6 & \textbf{33.4} & \textbf{76.3} & \textbf{41.6}   \\
		$\text{Ours}^{\dag}$  & \emph{val}  & ResNet-50-FPN &  \textbf{43.0} &  \textbf{79.0}  & \textbf{52.8}  & \textbf{50.5}  & \textbf{81.1 }& \textbf{61.5} & 31.8 & 75.9 & 39.4 \\
		\hline
		\hline
		AUNet~\cite{li2019attention} & \emph{test-dev} & ResNeXt-152-FPN & 46.5   & \textbf{81.0}  & 56.1   &  \textbf{55.9}   & \textbf{83.7} & \textbf{66.3} & 32.5 & 77.0 & 40.7  \\
		$\text{UpsNet}^{\dag}$~\cite{xiong2019upsnet} & \emph{test-dev}  &  DCN-101-FPN & 46.6   & 80.5   & 56.9  & 53.2 & 81.5 & 64.6 & \textbf{36.7} & \textbf{78.9} & \textbf{45.3}   \\
		$\text{Ours}^{\dag}$ & \emph{test-dev}  &  DCN-101-FPN & \textbf{47.3}   & 80.8   & \textbf{57.5}  & 54.9 & 82.1 & \textbf{66.3} & 35.9 & \textbf{78.9} & 44.3   \\
		\Xhline{3\arrayrulewidth}
	\end{tabular}
	\vspace{1em}
	\caption{\footnotesize Comparison with state-of-the-art methods on COCO \emph{val} and \emph{test-dev} set. $\dag$ refers to deformable convolution.}
	\label{tab:sota}
\end{table*}

\subsection{Implementation Details}
Our model is based on the implementation in~\cite{chen2019mmdetection}.  We extend the Mask-RCNN with a stuff head, and treat it as our baseline model. ResNet-50-FPN and DCN-101-FPN~\cite{dai2017deformable} are chosen as our backbone for \emph{val} and \emph{test-dev} respectively.

We use the SGD optimization algorithm with momentum of 0.9 and weight decay of 1e-4. 
For the model based on ResNet-50-FPN, we follow the 1x training schedule in~\cite{Detectron2018}. In the first 500 iterations, we adopt the linear warmup policy to increase the learning rate from 0.002 to 0.02. Then it is divided by 10 at 60k iterations and 80k iterations respectively. For the model based on DCN-101-FPN, we follow the 3x training schedule in~\cite{Detectron2018} and apply multi-scale training. \hl{The learning rate setting of the 3x schedule is adjusted in proportion to the 1x schedule.}
As for data augmentation, the shorter edge is resized to 800, while the longer side is kept below 1333. Random crop and horizontal flip are used. When training models containing I2S, we set $\lambda_s$ to 0.2 and $\lambda_i$ to $0.3$. For models without I2S, $\lambda_s$ is set to 0.5 since there is no $L_{I2S}$ left. For models that contain deformable convolutions, we set $\lambda_s$ to 0.1 and $\lambda_i$ to $0.2$. 

NMS is applied to all candidates whose scores are higher than 0.6 in a class-agnostic way, and its threshold is set to 0.5. In the occlusion handling algorithm, we first define the occluded pair as follows. For two objects $A$ and $B$, the pair $(A, B)$ is treated an occluded pair when the overlap area is larger than 20\% of either $A$ or $B$. 
When overlap ratio is less than 20\%, objects with higher scores simply overwrite the others. For all occluded pairs, we assign the overlapping part to the object with closer appearance as described in Section\mbox{~\ref{sec::texture}}. 
To handle the occlusion involving more than two objects, we deal with overlapping object pairs in descending order of pair scores, the higher object\mbox{’s} score in each pair. As for interweaving cases, where objects overlap each other, we would set aside the contradictory pairs with lower scores. For example, let $A\rightarrow B$ denotes that object A overlaps object B. Given $A\rightarrow B$, $C\rightarrow A$, $B\rightarrow C$ in an image with their pair scores in descending order, we would set $B\rightarrow C$ aside.
If more than 50\% of an object is assigned to other objects, we remove it from the scene.

After that, we resolve the conflicts between instances and stuff by prioritizing instances. Finally, we remove stuff regions whose areas are under 4096, as described in~\cite{kirillov2019panoptic1}.

\subsection{Comparison with State-of-the-Art Methods}
In Table~\ref{tab:sota}, we compare our method with other state-of-the-art methods~\cite{de2018panoptic} on COCO \emph{val} and \emph{test-dev} set. 

When comparing to methods without deformable convolution, our model outperforms them with respect to nearly all metrics on COCO \emph{val}. It achieves especially higher results at both SQ and RQ, showing that it is well-balanced between segmentation and recognition. 
By applying deformable convolutions in the network, our approach gains a clear improvement at PQ (from 41.1\% to \textbf{43.0\%}) and outperforms UpsNet on most of the metrics. When it comes to the performance on things, we achieved 50.5\% at $\text{PQ}^\text{Th}$ which exceeds UpsNet by 2\%. The improvement of $\text{PQ}^\text{Th}$ comes from having better $\text{SQ}^\text{Th}$(+1.6\%) and $\text{RQ}^{\text{Th}}$(+1.9\%). As for the performance on stuff, our method is inferior to UpsNet since we simply resolve the conflict between instances and segmentation in favor of instances. 

On COCO \emph{test-dev} set, our model based on DCN-101-FPN achieves a consistently higher performance of \textbf{47.3\%} PQ~(0.7\% higher than UPSNet).



\begin{figure*}[t]
	\centering
	\includegraphics[clip,width=0.95\textwidth]{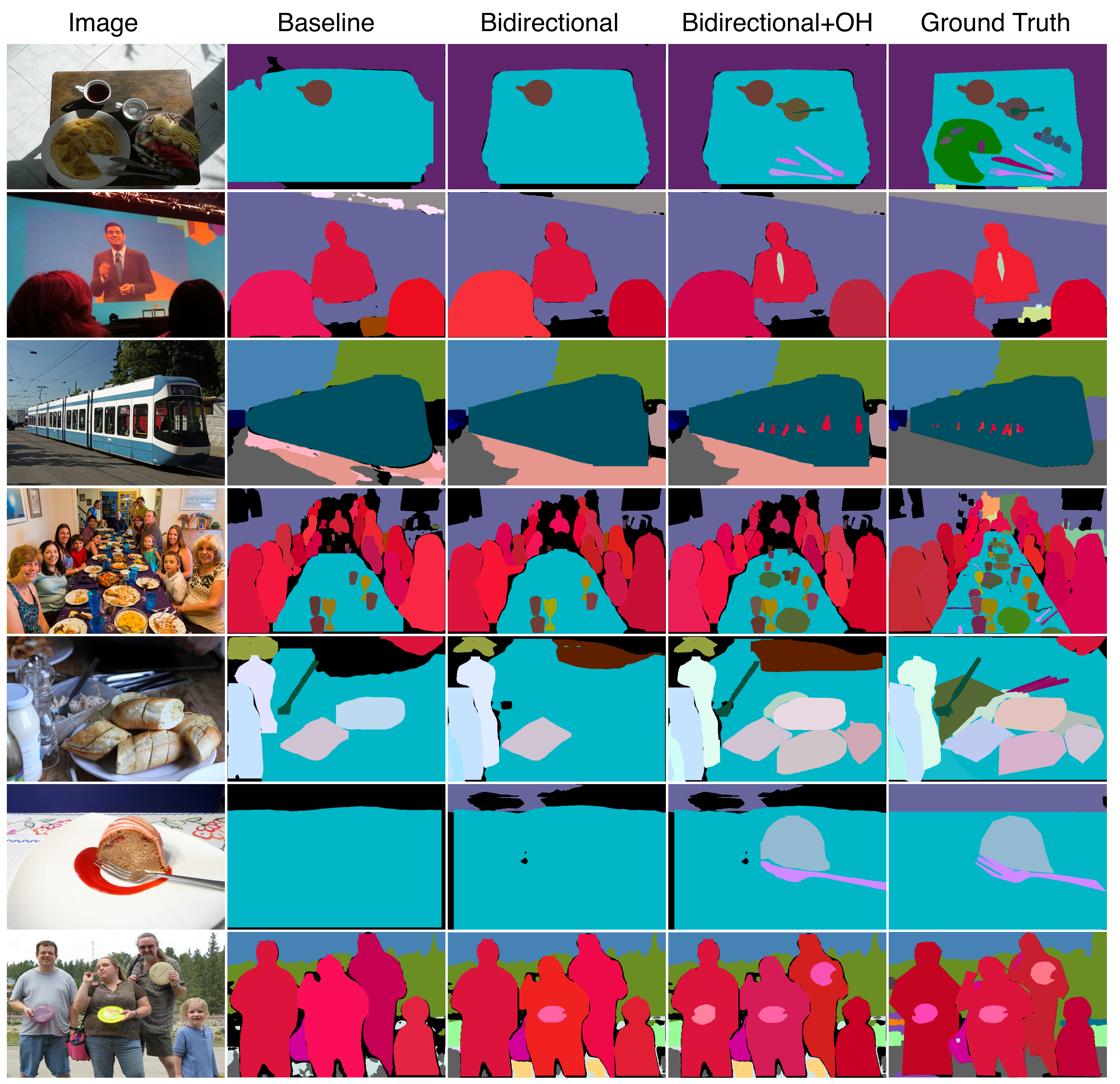}
	\caption{\footnotesize Visualization of panoptic segmentation results on COCO \emph{val}. ``Bidirectional'' refers to the combination of S2I and I2S. ``OH'' represents the occlusion handling module.
		The figure shows the improvements gained from our modules.}
	\label{fig:viz}
\end{figure*}
\subsection{Ablation Study}

We perform ablation studies on COCO \emph{val} with our model based on ResNet50-FPN. We study the effectiveness of our modules by adding them one-by-one to the baseline.

\paragraph{Instance-to-semantic}
To study the effect of Instance-To-Semantic~(I2S), we run experiments with SIM alone and with both SIM and OCM.  As shown in the second row of Table~\ref{tab:abl}, applying SIM alone leads to a 0.4\% gain in terms of PQ.  We notice that both $\text{SQ}^{\text{Th}}$ and $\text{SQ}^{\text{Sf}}$ get improved by more than 1\%. This demonstrates that SIM utilizes the recovered structural information to help semantic segmentation. Applying OCM together with SIM leads to another 0.5\% improvement in terms of PQ. Thanks to the object layout context provided by OCM, our model recognizes stuff regions better, resulting in 1.3\% improvement w.r.t. $\text{RQ}^\text{Sf}$.

\paragraph{Semantic-to-instance}
We apply S2I together with I2S, \ie, SIM and OCM. It turns out that S2I module can effectively improve $\text{RQ}^\text{Th}$(+0.4\%) by introducing complementary contextual information from semantic segmentation. The instance segmentation metric $\text{AP}_\text{mask}$ gets improved by 0.3\% as well. Although the semantic segmentation on stuff region~($\text{mIoU}^\text{Sf}$) maintains the same, $\text{PQ}^\text{Sf}$ is slightly improved by 0.2\% due to better thing predictions.
 
\paragraph{Deformable convolution}
To validate our modules' compatibility  with deformable convolution, we replace the vanilla convolution layers in the semantic head with deformable convolution layers. As shown in Table~\ref{tab:abl}, deformable convolution improves our model's performance by 1.5\% and is extremely helpful for ``stuff'' regions, as evidenced by the 1.3\% increment of $\text{PQ}^\text{Sf}$.

\begin{table*}[tb]
	\small
	\centering
	\begin{tabular}{c c c c c | c c c | c c c |  c c c | c c }
		\Xhline{3\arrayrulewidth}
		SIM & OCM & S2I & DFM & OH & $\text{PQ}$ & $\text{SQ}$ & $\text{RQ}$ & 
		$\text{PQ}^\text{Th}$ & $\text{SQ}^\text{Th}$ &   $\text{RQ}^\text{Th}$ & 
		$\text{PQ}^\text{Sf}$ & $\text{SQ}^\text{Sf}$ &   $\text{RQ}^\text{Sf}$ & 
		$\text{AP}_\text{mask}$ & $\text{mIoU}^{\text{Sf}}$ \\
		\hline
		\hline
		& & & & & 39.1 & 77.3 & 48.1 & 46.7 & 80.4 & 56.6 & 27.7 & 72.5 & 35.4 & 34.2 & 38.6 \\
		\checkmark & & & & & 39.5 & 78.0 & 48.6 & 47.1 & 81.2 & 57.1 & 28.0 & 73.1 & 35.8 & 34.6 & 39.5 \\
		\checkmark & \checkmark &  & & & 40.0 & 78.4 & 49.1 & 47.2 & 81.6 & 57.0 & 29.2 & 73.5 & 37.1 & 34.8 & 39.7 \\
		\checkmark & \checkmark & \checkmark & & & 40.3 & 78.1 & 49.5 & 47.5 & 81.5 & 57.4 & 29.4 & 73.2 & 37.3 & 35.1 & 39.7\\
		\checkmark & \checkmark & \checkmark & \checkmark & & 41.8 & 79.6 & 50.8 & 48.5 & 82.1 & 58.3 & 31.7 & 75.9 & 39.4 & 36.4 & 41.1 \\
		\checkmark & \checkmark & \checkmark & \checkmark & \checkmark & 43.0 & 79.0 & 52.8 & 50.5 & 81.1 & 61.5 & 31.8 & 75.9 & 39.5 & 36.4 & 41.1\\
		\Xhline{3\arrayrulewidth}
	\end{tabular}
	\vspace{0.93em}
	\caption{\footnotesize Ablation study on COCO \emph{val}. `SIM', `OCM'  are modules used in Instance-To-Semantic. S2I stands for Semantic-To-Instance. DFM stands for deformable convolution. OH refers to the occlusion handling algorithm. All results without OH are obtained by the heuristic fusion~\cite{kirillov2019panoptic1}.}
	\label{tab:abl}
\end{table*}

\paragraph{Occlusion handling}
Occlusion handling is aimed at resolving occlusion between object instances and assigning occluded pixels to the correct object. Our occlusion handler makes use of local appearance~(RGB) information and is completely learning-free. By applying the proposed occlusion handling algorithm, we greatly improve the recognition of things, as reflected by a 2\% increase w.r.t. $\text{PQ}^{\text{Th}}$. Due to the better object arrangement provided by our algorithm, $\text{PQ}^{\text{Sf}}$ is also slightly improved~(+0.1\%).

\begin{table}[tb]
	\small
	\centering
	\begin{tabular}{c c  | c c c }
		\Xhline{3\arrayrulewidth}
		Backbone & OH & $\text{PQ}$ & $\text{PQ}^\text{Th}$ & $\text{PQ}^\text{Sf}$ \\
		\hline
		\hline
		ResNet-50-FPN &  & 41.8 & 48.5 & 31.7\\
		ResNet-101-FPN &  & 42.5 & 48.6 & 33.4\\
		\hline
		ResNet-50-FPN & \checkmark & 43.0 & 50.5 & 31.8 \\
		ResNet-101-FPN & \checkmark & 44.0 & 51.0 & 33.4\\
		\Xhline{3\arrayrulewidth}
	\end{tabular}
	\vspace{0.93em}
	\caption{\footnotesize Experimental results for our method with different backbones.}
	\label{tab:backbone}
\end{table}
\paragraph{Different backbones}
We analyze the effect of the backbone by comparing different backbone networks. The performance of our model can be further improved to $44.0\%$ by adopting a deeper ResNet-101-FPN backbone. As shown in Table~\ref{tab:backbone}, without the occlusion handling algorithm, the model based on ResNet-101-FPN is 0.8\% higher than ResNet-50-FPN. When both applying the occlusion handling algorithm, our model based on ResNet-101-FPN achieves 1.0\% better performance than ResNet-50-FPN. This also reveals that our occlusion handling algorithm can improve $\text{PQ}^{\text{Th}}$ consistently based on different backbones.

\paragraph{Bottleneck analysis}
\begin{table}[tb]
	\small
	\centering
	\begin{tabular}{c c c c | c c c }
		\Xhline{3\arrayrulewidth}
		\tabincell{c}{GT \\ Box} & \tabincell{c}{GT \\ ICA} & \tabincell{c}{GT
 			\\ Occ} & \tabincell{c}{GT \\ Seg} & $\text{PQ}$ & $\text{PQ}^\text{Th}$ & $\text{PQ}^\text{Sf}$ \\
		\hline
		\hline
		& & & & 43.0 & 50.5 & 31.8\\
		& & \checkmark & & 44.6 & 53.2 & 31.8\\
		\checkmark & & & & 47.1 & 56.6 & 32.8\\
		\checkmark & \checkmark & & & 58.4 & 74.8 & 33.5\\
		\checkmark & \checkmark & \checkmark  & & 59.3 & 76.3 & 33.5 \\
		& & & \checkmark & 60.8 &50.5 & 76.4 \\
		\Xhline{3\arrayrulewidth}
	\end{tabular}
	\vspace{0.93em}
	\caption{\footnotesize Bottleneck analysis on COCO \emph{val}. We feed different types of ground truth into our model. GT Box stands for ground truth boxes. GT ICA refers to assigning the ground truth classes to instances. GT Occ means the ground truth overlap relationship. GT Seg denotes ground truth semantic segmentation.}
	\label{tab:bottle}
\end{table}

To analyze the performance bottleneck of our approach, we replace parts of the intermediate results with the ground truth to see how much improvement it will lead to. Specifically, we study ground truth overlap relationships, ground truth boxes, ground truth instance class assignment and ground truth segmentation as input. 

To estimate the potential of the occlusion algorithm, we feed ground truth overlaps into the model. Specifically, the predicted boxes are first matched with ground truth boxes. Then the occlusion among matched predictions is resolved using ground truth overlap relationship. The rest of the unmatched occluded predictions are handled by our occlusion handling algorithm. As shown in Table~\ref{tab:bottle}, when feeding ground truth overlaps, the performance $\text{PQ}^\text{Th}$ increases to 53.2\%. This demonstrates that there still exists a large gap between our occlusion algorithm and an ideal one.

By feeding ground truth boxes, PQ for both things and stuff sees an increase of 6.1\% and 1\% respectively, which indicates the maximum performance gain of a better RPN. We further assign the predictions of boxes to ground truth labels, which increases $\text{PQ}^\text{Th}$ by more than 20\%. This demonstrates that the lack of recognition ability on things is a main bottleneck of our model. Meanwhile, We also test feeding ground truth overlap along with ground truth box and class assignment,  $\text{PQ}^\text{Th}$ gets a further improvement of 2\%. This shows that the occlusion problem has to be carefully dealt with even if ground truth boxes and labels are fed. Finally, we test the case when ground truth segmentation is given,  the performance of $\text{PQ}^\text{Sf}$ is only 76.4\%. This indicates that the common fusion process that prioritizes things over stuff is far from optimal.

\paragraph{Visualization}
We show visual examples of the results obtained by our method in Figure~\ref{fig:viz}. By comparing the second and third columns, we can see large improvements brought by using the bidirectional architecture, specifically, many large misclassified regions are corrected. After adding the occlusion handling module~(fourth column) we notice that several conflicts of instances are resolved. This causes the accuracy of overlapping objects to increase significantly.

\section{Conclusion}
In this paper, we show that our proposed bidirectional learning architecture for panoptic segmentation is able to effectively utilize both instance and semantic features in a complementary fashion. Additionally, we use our occlusion handling module to demonstrate the importance of low-level appearance features for resolving the pixel to instance assignment problem. The proposed approach achieves the state-of-the-art result and the effectiveness of each of our modules is validated in the experiments.

\paragraph{Acknowledgment}
This work is in part supported by key scientific technological innovation research project by Ministry of Education, Zhejiang Provincial Natural Science Foundation of China under Grant LR19F020004, Baidu AI Frontier Technology Joint Research Program, Zhejiang University K.P. Chao's High Technology Development Foundation. 

\newpage
{\small
\bibliographystyle{ieee_fullname}
\bibliography{egbib}
}
\clearpage 
\appendix
\noindent\textbf{\Large Appendices}

\section{Different Patch-Recovering Operators}

In this section, we compare three patch-recovering operators, \ie, RoIInlay, RoIUpsample and Avg RoIUpsample~(a modified version of RoIUpsample) from three aspects, namely visual effect, runtime and experiment results.

\begin{figure}[h]
	\centering
	\includegraphics[width=0.48\textwidth]{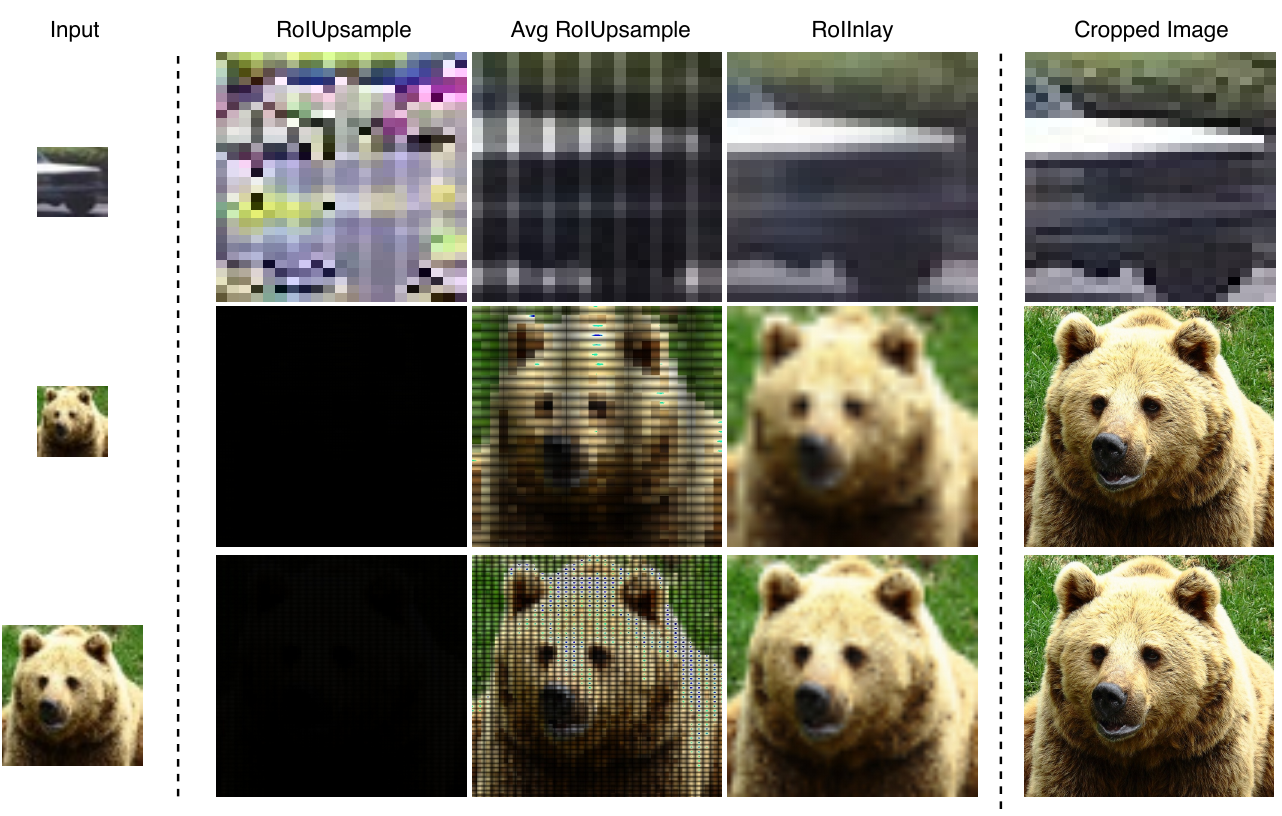}
	\caption{\footnotesize Comparison of RoIInlay, RoIUpsample and its modified version Avg RoIUpsample. All of them take the output of RoIAlign as input. The output size of RoIAlign is set to $28 \times 28$ for the first and second row and $56 \times 56$ for the third row. The object in the first row is small~($<28 \times 28$), while the one in the second and third row is large~($>56 \times 56$).}
	\label{fig:op}
\end{figure}
\paragraph{Visual effect}
A better patch-recovering operator has an output that looks more similar to the original cropped image.  
To give a visual comparison between RoIInlay and RoIUpsample, we test them on raw images~(RGB). Specifically, we first use RoIAlign to obtain a resized patch of an object, and then feed it to both operators to recover its resolution. Moreover, a modified version of RoIUpsample, named as Avg RoIUpsample, is provided. It replaces the summation in RoIUpsample with averaging. As shown in Figure~\ref{fig:op}, our RoIInlay performs better no matter what object area and what output setting of RoIAlign is, while either RoIUpsample or its modified version suffers from black stripes. These black stripes are formed by sampling ``holes'' of RoIUpsample, as described before. The output of RoIUpsample is extremely different since it sums values from multiple reference points, leading to the change of the scale of values.

\begin{table}[h]
	\scriptsize
	\centering
	\begin{tabular}{c c c c c c}
		\Xhline{3\arrayrulewidth}
		\#Objects & \tabincell{c}{Object \\size} & \tabincell{c}{Output \\size} & \tabincell{c}{RoI \\ Upsample}~(ms) & \tabincell{c}{RoI \\ Inaly}~(ms) & Speed-up \\ 
		\hline
		50 & $28$ & $300$ & 3.65 & 2.17 & $\times1.68$ \\ 
		100 & $28$ & $300$ & 6.8 & 3.9 & $\times1.74$\\ 
		100 & $28$ & $800$ & 9.7 & 6.7 & $\times1.45$ \\ 
		100 & $56$ & $300$ & 34.9 & 12.1 & $\times2.88$ \\ 
		100 & $128$ & $300$ & 440.5 & 122.4 & $\times 3.60$ \\ 
		\Xhline{3\arrayrulewidth}
	\end{tabular}
	\vspace{1em}
	\caption{\footnotesize Speed comparison between RoIInlay and RoIUpsample on GTX 1080Ti. The input of both operators are tensors of 512 channels. They are resized according to object sizes and put into a tensor of output size as output.}
	\label{tab:speed}
\end{table}
\paragraph{Runtime}
To test the speed of RoIInlay,  we record its execution time and compare it with RoIUpsample's on GTX 1080Ti. As shown in Table~\ref{tab:speed}, RoIInlay is faster than RoIUpsample under various settings. The speedup ratio grows to $\times 3.60$ when object size is set to 128. In the COCO dataset, the average object size is $98 \times 98$, indicating that we can obtain about $\times3$ speedup with RoIInlay.

\paragraph{Experimental results}
We test these two operators on the model with only SIM module to show how it affects the actual performance. As shown in Table~\ref{tab:Exp}, RoIUpsample will hurt the segmentation quality~(SQ) by about \textbf{1\%}. 

\begin{table}[h]
	\small
	\centering
	\begin{tabular}{c c c c }
		\Xhline{3\arrayrulewidth}
		Operator & PQ & SQ & RQ  \\
		\hline
		RoIUpsample & 39.4 & 77.0& \textbf{48.7} \\
		RoIInlay & \textbf{39.5} & \textbf{78.0} & \textbf{48.7} \\
		\Xhline{3\arrayrulewidth}
	\end{tabular}
	\vspace{1em}
	\caption{\footnotesize Experimental results on COCO \emph{val} for RoIInlay and RoIUpsample. Both operators are applied to the model with only SIM module. The random seed and training schedule is set to the same.}
	\label{tab:Exp}
\end{table}

\end{document}